\documentclass{article}

\usepackage[preprint]{neurips_2023}

\usepackage{tabularx} 

\usepackage[utf8]{inputenc} 
\usepackage[T1]{fontenc}    
\usepackage{hyperref}       
\usepackage{url}            
\usepackage{booktabs}       
\usepackage{amsfonts}       
\usepackage{nicefrac}       
\usepackage{microtype}      
\usepackage{xcolor}         
\usepackage{graphicx}
\usepackage{algorithm2e}
\usepackage{svg}

\title{Brain Captioning: Decoding human brain activity into images and text}

\author{%
Matteo Ferrante \\ Department of Biomedicine and Prevention \\ University of Rome Tor Vergata (IT)
\And 
Furkan Ozcelik \\ CerCo, CNRS UMR5549, Toulouse, France \\ Universite de Toulouse, Toulouse, France
\And 
Tommaso Boccato \\Department of Biomedicine and Prevention \\ University of Rome Tor Vergata (IT)
\AND 
Rufin VanRullen \\ CerCo, CNRS UMR5549, Toulouse, France \\ Universite de Toulouse, Toulouse, France \\ ANITI, Toulouse, France
\And 
Nicola Toschi \\ Department of Biomedicine and Prevention \\ University of Rome Tor Vergata (IT) \\ Martinos Center For Biomedical Imaging \\ MGH and Harvard Medical School (USA)}

\begin{document}

\maketitle



\begin{abstract}

Every day, the human brain processes an immense volume of visual information, relying on intricate neural mechanisms to perceive and interpret these stimuli. Recent breakthroughs in functional magnetic resonance imaging (fMRI) have enabled scientists to extract visual information from human brain activity patterns. In this study, we present an innovative method for decoding brain activity into meaningful images and captions, with a specific focus on brain captioning due to its enhanced flexibility as compared to brain decoding into images. Our approach takes advantage of cutting-edge image captioning models and incorporates a unique image reconstruction pipeline that utilizes latent diffusion models and depth estimation. \\
We utilized the Natural Scenes Dataset, a comprehensive fMRI dataset from eight subjects who viewed images from the COCO dataset. We employed the Generative Image-to-text Transformer (GIT) as our backbone for captioning and propose a new image reconstruction pipeline based on latent diffusion models. The method involves training regularized linear regression models between brain activity and extracted features. Additionally, we incorporated depth maps from the ControlNet model to further guide the reconstruction process.\\
We evaluate our methods using quantitative metrics for both generated captions and images. Our brain captioning approach outperforms existing methods, while our image reconstruction pipeline generates plausible images with improved spatial relationships. \\
In conclusion, we demonstrate significant progress in brain decoding, showcasing the enormous potential of integrating vision and language to better understand human cognition. Our approach provides a flexible platform for future research, with potential applications in various fields, including neural art, style transfer, and portable devices.

\end{abstract}

\begin{figure}[h!]
\includegraphics[width=.9\textwidth]{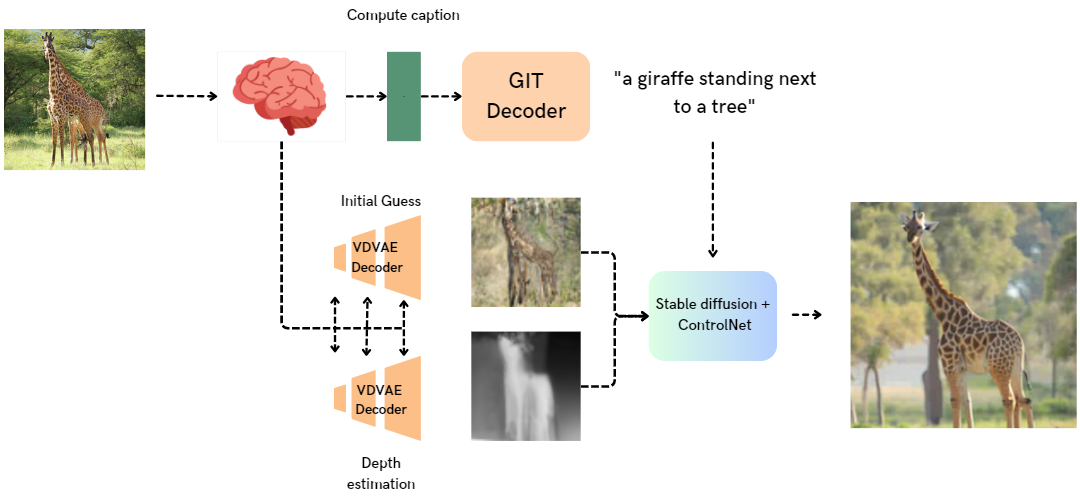}
\caption{Our model utilizes fMRI measurements to extract features for GIT captioning and VDVAE initial and depth image estimation using linear models. Image captions serve as the primary general result, used in the second stage alongside other conditioning to generate plausible reconstructions with a latent diffusion model. GIT and VDVAE models are pre-trained and frozen, while linear regressions are trained from fMRI to their latent spaces.}
\label{fig:whole_pipeline}
\end{figure}
\section{Introduction}

\begin{figure}[h!]
\centering
\includegraphics[width=0.95\textwidth]{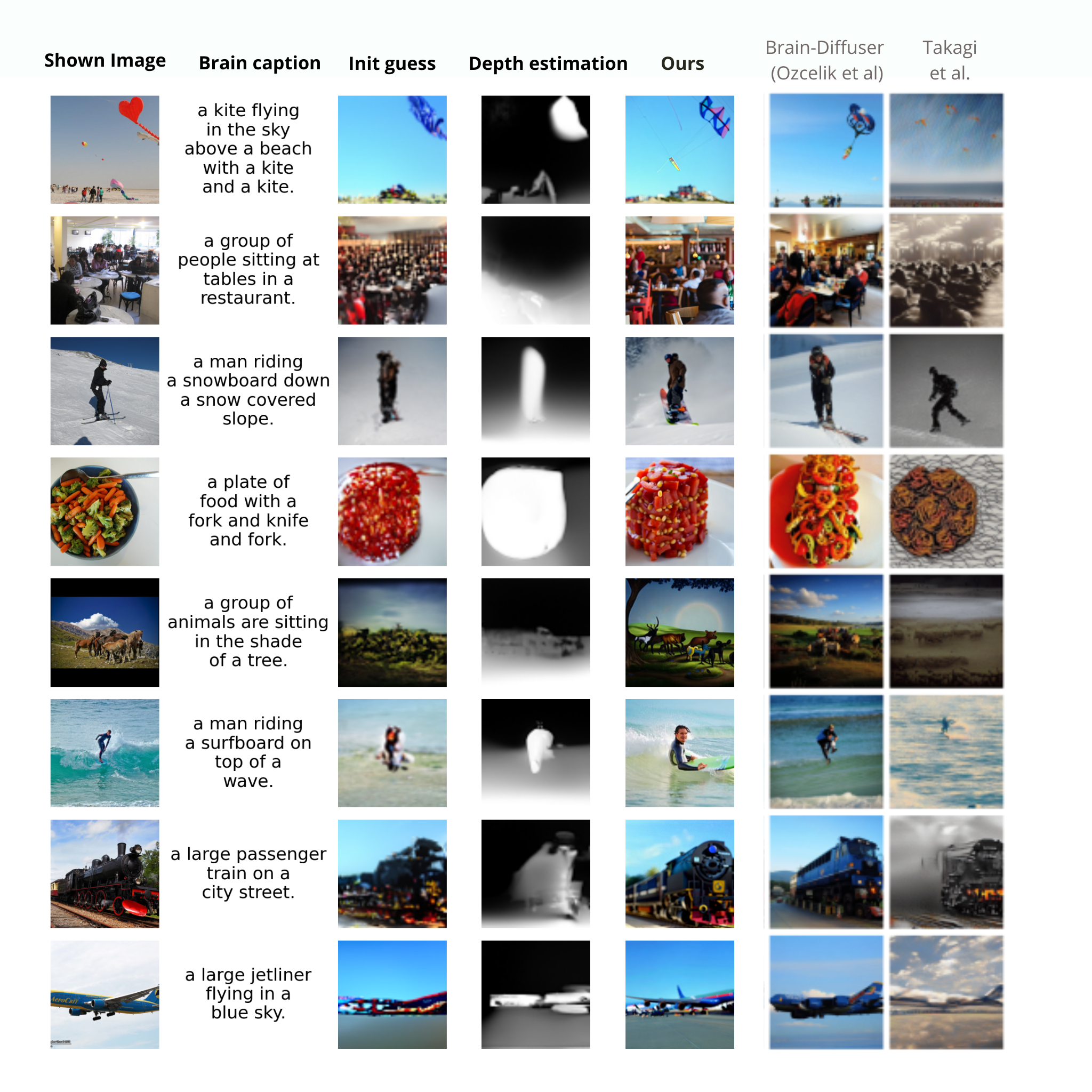}
\caption{Comparison of our results (Columns 2-4) with the shown stimuli and reconstructions from other works. The second column displays the caption computed from the brain activity, the third column presents the initial guess image, the fourth column shows the depth estimated images, and the fifth column reports our final reconstruction. The last two columns showcase reconstructions from two recent works. All results are from subj01.}
\label{fig:compare_takagi}
\end{figure}

The human visual system is an extraordinary product of evolution, enabling us to navigate and interact with our surroundings. From basic patterns to intricate scenes, our brains persistently process and interpret visual information. A central challenge in neuroscience is comprehending how these elaborate processes occur at the neural activity level.

Functional magnetic resonance imaging (fMRI) has emerged as an essential tool for studying neural activity associated with visual perception, by measuring blood oxygen level-dependent (BOLD) signals. Brain decoding has progressed significantly, employing fMRI data to reconstruct visual stimuli from brain activity patterns. This has the potential to revolutionize our understanding of the neural code underlying visual perception with possible applications in brain-computer interfaces and clinical diagnostics. The increasing interest in reconstructing information from noninvasive brain data is driven by enhanced data availability, improved computational power, and sophisticated deep learning methods. Despite challenges with signal-to-noise ratio, session duration, and hemodynamic response function variability, fMRI has proven effective in various tasks such as visual stimulus and text classification and reconstruction \citep{CEBRA,zafar_decoding_2015,lindsay_convolutional_2021,awangga_literature_2020}.

In this work, our first contribution is shifting the prediction from images to text, aiming to generate a caption of the observed scene from brain activity. To compare with prior work, we propose a new model for image captioning from brain activity and propose a new image reconstruction pipeline based on a conditioned and controlled version of the latent diffusion model, Stable Diffusion. Predicting a caption instead of the image in brain decoding from fMRI of visual stimuli offers several advantages. Captions naturally represent a higher level of abstraction, requiring a more advanced interpretation and summarization of visual information than merely predicting the image itself. As a result, predicting captions can help us understand how the brain processes and represents complex visual information. In real-world situations, humans often describe visual scenes with words, so predicting captions instead of images may better capture an important aspect of visual information processing. Recent neuroscience research has shown substantial evidence that large language models can be correlated with brain activity and that it is possible to predict one representation from the other \citep{convergence_language,tang_semantic_2023_semantic_language}. 
Finally, predicting text from fMRI could lead to better generalization across modalities. Natural language is our main tool as humans to interact with each other and nowadays even with foundation models. We can exploit large language models to condition other models to generate images, videos, audio, and more. Predicting text from brain helps us rapidly change the reconstruction model, leveraging state-of-the-art text-to-image models to generate realistic images from brain activity.

In summary, our contributions in this paper are two-fold:
We propose a method to generate image captions from brain activity using a multimodal large language model \citep{GIT} and introduce a novel image reconstruction pipeline based on predicted text and estimated initial and depth maps from brain activity. Fig \ref{fig:whole_pipeline} is a scheme of the entire procedure that we propose, while Fig \ref{fig:compare_takagi} shows generated captions and images from brain activity compared to other image reconstruction methods.

\subsection{Related Works}

In the field of brain decoding, researchers have utilized various modeling frameworks with preprocessed fMRI time series as input. These data have served as the basis for numerous decoding approaches. Some examples include employing a variational autoencoder with a generative adversarial component (VAE-GAN) to encode latent representations of human faces \citep{faces} and applying sparse linear regression on preprocessed fMRI data to predict features extracted from early convolutional layers in a pre-trained CNN \citep{horikawa_generic_2017} for natural images. Unsupervised and adversarial strategies have been used to reconstruct images, incorporating dual VAE-GAN and unsupervised methods for fMRI stimuli decoding with various encoders and decoders trained in different ways \citep{shen_end--end_nodate,ren_reconstructing_2019,gaziv_self-supervised_2022}. Optimizing the latent spaces of pretrained architectures, such as BigBiGAN and IC-GAN, can facilitate reconstructing high-quality images from fMRI patterns \citep{bigbigan,icgan,mozafari_reconstructing_2020,ozcelik_reconstruction_2022}. Recently, diffusion models have become a significant component of the decoding pipeline due to their improved performance in image generation \citep{Takagi2022.11.18.517004, chen2022seeing}, also incorporating semantic-based strategies like \citep{ferrante2023semantic} or multi-step decoding strategies as in \citep{ozcelik2023braindiffuser}. To the best of our knowledge, only a few works \citep{takada_generation_2020,matsuo_generating_2016,qiao_accurate_2018} have attempted brain captioning, utilizing a combination of a pre-trained convolutional neural network and recurrent neural network for captioning and estimating the convolutional features from brain activity. The primary differences between our work and previous research are the shift in paradigm from direct image estimation to brain captioning and leveraging multimodal transformer-based language models, which have been shown to better describe brain activity \citep{choksi_multimodal_2022}.

\section{Methods}
\begin{figure}[h!]
\centering
\includegraphics[width=0.9\textwidth]{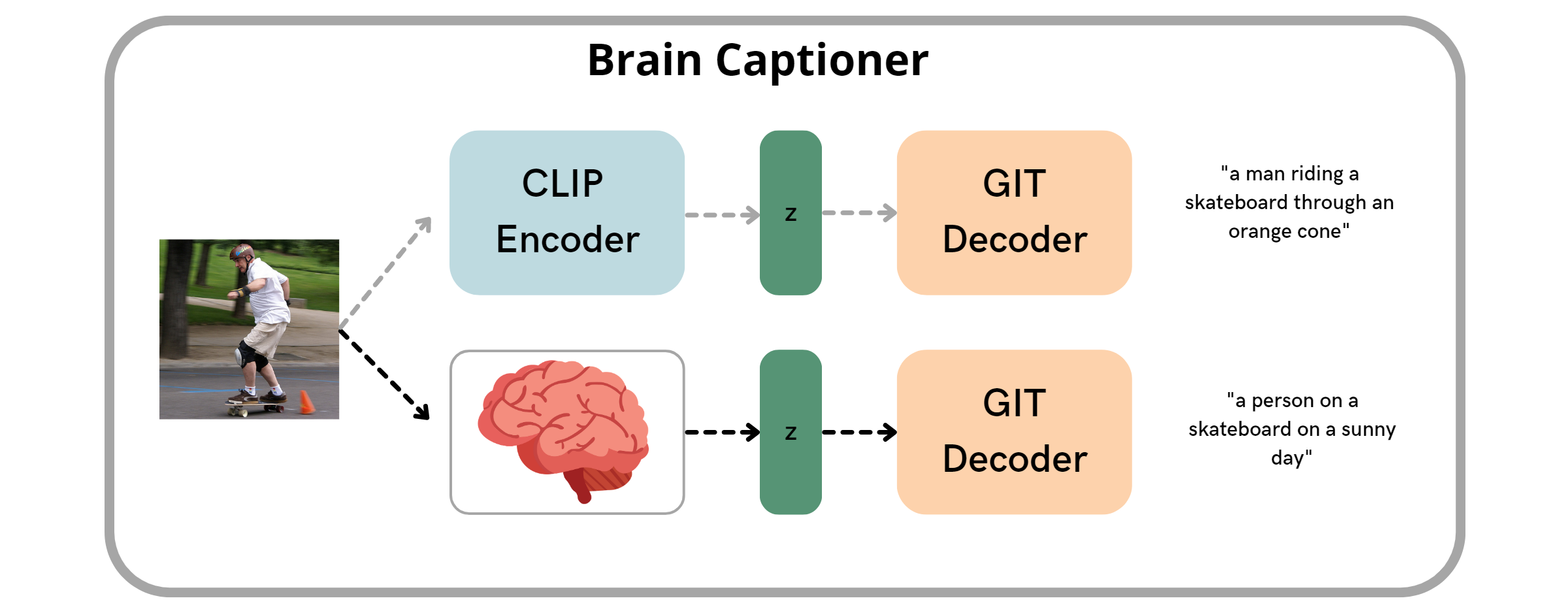}
\caption{Image captioning from brain activity pipeline: Gray dotted lines are only used during training, and only orange boxes are used during inference, replacing their inputs with those estimated from brain activity.}
\label{fig:captioning_pipeline}
\end{figure}

In this section, we describe the proposed method and the data we used. The data are publicly available and can be requested at \url{https://naturalscenesdataset.org/}. All experiments and models were trained on a server equipped with four A100 GPU cards and 2 TB of RAM. The entire analysis took approximately 16 hours per subject. The pipelines are based on pre-trained versions of deep learning models used as proxies for brain activity, generating latent representations that could be similar (and thus linearly mapped) to brain activity and vice versa.

\begin{figure}
    \centering
    \includegraphics[width=\textwidth]{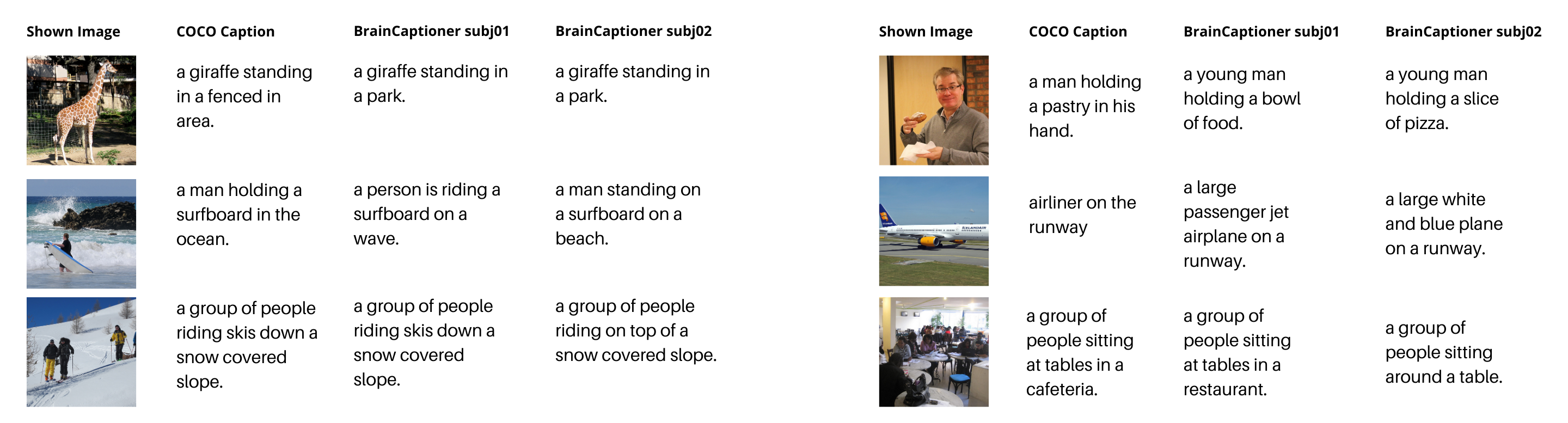}
    \caption{Examples of generated caption with our BrainCaptioner pipeline. Shown images are test set stimuli used for subj01 and subj02 during the fmri experiment. COCO Caption column report the first annotations for the original COCO image, while the other two columns are the output of our model for the two subjects.}
    \label{fig:captions}
\end{figure}

\subsection{Data}
We employed the Natural Scenes Dataset (NSD) \citep{NSDDataset}, a comprehensive fMRI dataset featuring eight subjects who viewed images from the COCO dataset. Our analysis concentrated on two subjects (subj01 and subj02), yielding a training set of 8,859 images and 24,980 fMRI trials, and a test set of 982 images and 2,770 fMRI trials per subject. Images are repeated up to three times and their trials were averaged to increase signal-to-noise ratio.  To reduce spatial dimensionality to approximately 15,000 voxels, the fMRI signal (1.8mm resolution) was masked using the NSDGeneral ROI mask, which covers numerous visual areas. This ROI selection is vital for enhancing the signal-to-noise ratio and minimizing data complexity. The chosen ROI mask facilitated the investigation of both low-level and high-level visual features. To decrease temporal dimensionality, we employed precomputed betas from a GLM with fitted HRF and denoised as described in the NSD paper.

\subsection{Captioning model and renormalization}

For brain captioning, we utilized the state-of-the-art image captioning model, GIT \citep{GIT}, as our backbone. GIT (Generative Image-to-text Transformer) is an innovative model designed to integrate vision and language tasks. In contrast to conventional approaches that depend on intricate architectures and external modules, GIT adopts a streamlined structure consisting of a single image encoder and a text decoder, unified under one language modeling task. Leveraging large-scale pre-training data and model size, GIT outperforms existing models on 12 benchmarks and even surpasses human performance on TextCaps. Essentially, GIT comprises a CLIP Vision encoder \citep{clip} followed by a GPT decoder, trained on large-scale datasets.

For the stimuli in the train set, we computed features from images and trained a regularized linear regression to map between brain activity and these features. We used cross-validation to select the best regularization parameter $\alpha$ and discovered that a value of 50,000 performed optimally using the negative mean squared error as a scoring function. This is our brain-to-features model, which serves as the core component of our method for brain captioning.

Before feeding estimated features to the decoder, we required a normalization pass. Thus, we computed the mean and standard deviation of features from images and those predicted by the model over the training set, replacing their values during inference on the test set to match the real feature distributions. A schematic representation of the overall pipeline can be seen in Fig. \ref{fig:captioning_pipeline} and generated captions from this pipeline for both subjects are  shown in Fig \ref{fig:captions}.


\subsection{Image reconstruction pipeline}

Recent research in brain decoding has focused on developing image reconstruction techniques \citep{ozcelik2023braindiffuser,ozcelik_reconstruction_2022,Takagi2022.11.18.517004,lin2022mind,chen2022seeing}. Studies have demonstrated that high SNR fMRI data of visual stimuli enables effective brain decoding using diffusion models. Various approaches have been proposed to enhance these models' performance, with the optimal method for image reconstruction remaining an open question. One approach to improve low-level detail generation and increase the similarity between original and decoded images is to provide the network with an initial guess image or an estimated latent space.

\begin{figure}[!h]
\centering
\includegraphics[width=1\textwidth]{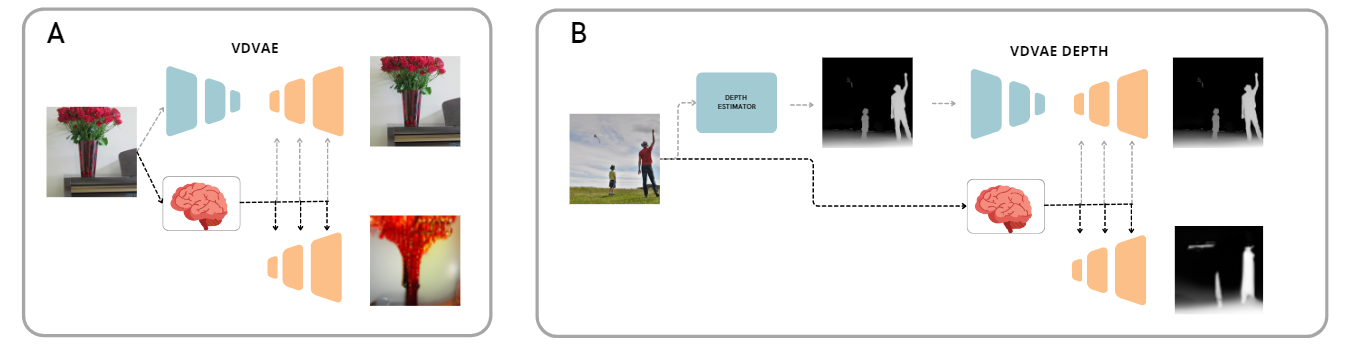}
\caption{\textbf{A}: Pipeline for initial images capturing 2D RGB pixel information. \textbf{B}: Pipeline for inferred depth estimates.
Both depth image and the initial image are estimated from brain activity. Gray dotted lines are only used during training, while only orange boxes are used during inference, replacing their inputs with the ones estimated from brain activity.}
\label{fig:depth}
\end{figure}

\textbf{Initial Guess:}

To compare our approach with existing research on brain decoding, we augmented our method by proposing an image reconstruction pipeline based on latent diffusion models.

Following the approach described in \citep{ozcelik2023braindiffuser}, we initially estimate a "guess image" to generate an approximate initial image with colors and shapes. To achieve this, we computed the latent representations of the first 31 layers of the very deep variational autoencoder model \citep{vdvae} (VDVAE), pre-trained on natural images, and kept frozen.

In a VDVAE, the encoder network maps the input data onto a lower-dimensional latent space, while the decoder network maps the latent space back to the original data space. The architecture of the VAE is hierarchical, featuring multiple layers of hidden units that extract increasingly abstract features from the input data. The most critical aspect of this is the conditional dependence between the hierarchical layers. In other words, the hidden units in each layer depend not only on the input data but also on the outputs of the previous layer. This conditional dependence allows the VAE to capture complex relationships between the input data and the latent space, resulting in a more powerful and expressive model. Consequently, we trained a regularized linear regression between brain activity and estimated features for each of the first 31 layers, using the renormalization procedure described in the previous section to match the target distribution.

During inference over the test set, these features are estimated from brain activity, renormalized, and passed to the VDVAE decoder to reconstruct an initial image, as depicted in Fig. \ref{fig:depth}.

\textbf{Depth estimation}:

We propose using ControlNet \citep{controlnet} to augment Stable Diffusion \citep{stable_diffusion}, a state-of-the-art latent diffusion model, for improving foreground-background matching in reconstructed images by incorporating depth information. We first compute grayscale depth images for all training stimuli using Dense Vision Transformer and the Huggingface library \citep{dpt,huggingface}. We then pass these depth images into the Variational Diffusion Autoencoder (VDVAE) model and train a regularized linear regression from brain activity to the model's latent, as illustrated in Fig \ref{fig:depth}.The VDVAE is the same used before (pre-trained on natural images and kept frozen), however here it is here to generate latent representation of the estimated depth images, which are our target for regression. 

\textbf{Whole Reconstruction pipeline}: 

The pipeline (Fig \ref{fig:whole_pipeline}) first decodes brain activity into a latent space to generate captions for test stimuli using learned ridge regression. Then, the initial guess and depth images are computed from brain activity to condition the latent model. Stable Diffusion v2 + ControlNet is used for implementation, with 30 inference steps, guidance scale 9, and control net weight 0.8. The negative prompt sentence \textit is also included to improve quality.

\subsection{Evaluation}
We compared our brain captioning work with existing methods by re-implementing the architecture from \citep{takada_generation_2020}, consisting of a CNN followed by an LSTM. We used Ridge regression to map brain activity to the CNN's final convolutional layer and applied renormalization before feeding the LSTM. We evaluated the generated captions using metrics such as  METEOR, CLIP similarity, and SentenceTransformer similarity. Additionally, we assessed our image reconstruction pipeline using low-level and high-level metrics like PixCorr, SSIM, 2-way accuracy in AlexNet, Inception, and CLIP latent spaces, and FID, allowing comparison with other brain decoding studies.

\section{Results}

\begin{table}[]
\resizebox{\textwidth}{!}{\begin{tabular}{lllll}
\textbf{Metric}                & \textbf{baseline subj01} & \textbf{baseline subj02} & \textbf{Ours (subj01)} & \textbf{Ours (subj02)} \\
Meteor (image captions and human captions)  & 0,176                    & 0,174                    & \textbf{0,404}         & \textbf{0,404}         \\
\textbf{Meteor (brain captions and image captions)}   & 0,163                    & 0,166                    & \textbf{0,305}         & \textbf{0,298}         \\
Sentence  (image captions and human captions)            & 0,319                    & 0,315                    & \textbf{0,703}         & \textbf{0,703}         \\
\textbf{Sentence (brain captions and image captions)} & 0,280                    & 0,281                    & \textbf{0,447}         & \textbf{0,418}         \\
CLIP  (image captions and human captions)               & 0,672                    & 0,673                    & \textbf{0,831}         & \textbf{0,831}         \\
\textbf{CLIP (brain captions and image captions)}     & 0,624                    & 0,627                    & \textbf{0,705}         & \textbf{0,688}        
\end{tabular}}
\caption{Text metrics: Here are reported the values of the metrics for each subject both for baseline and our model (columns). Each row is a different metric. Metrics with "(image captions and human captions)" are the evaluation of the model-generated captions from images versus the original COCO captions. These serve as a comparison of the model performances, while metrics with "(brain captions and image captions)" are metrics relative to captions computed from brain activity. }
\label{tab:text}
\end{table}
Table \ref{tab:text} presents the results of the evaluation of the proposed approach compared to the baseline models and previous works. This table reports text-based metrics, including Meteor score, CLIP, and SentenceTransformer similarity, computed for the reference captions, captions generated from images by both models (baseline and proposed), and captions generated from brain activity using the proposed approach. Results show that our approach outperforms the baseline models on all metrics and achieves significantly higher scores than previous works, indicating the effectiveness of the approach in generating accurate and meaningful captions from brain activity.

The table \ref{tab:img} reports image-based metrics, including PixCorr, SSIM, accuracy in various layers of AlexNet and Inception, CLIP similarity, and FID score. Results show that the proposed approach outperforms the previous works in low-level metrics, including PixCorr, SSIM and the lower layer of AlexNet. High level metrics are on par or slightly lower than state-of-the-art methods, probably due to a bottleneck in text predictions. If a word is predicted wrongly, this error is propagated in the image reconstruction pipeline and impacts on high-level metrics. Overall, the results demonstrate the effectiveness of the proposed approach in decoding brain activity into meaningful images and captions, performing on par on even outperforming state-of-the-art in several metrics.
Fig \ref{fig:compare_takagi}, \ref{fig:compare_gu},  \ref{fig:captions} and figures in the supplementary material show some visual comparison with other works for a qualitative comparison. Qualitatively, the captions represent plausible descriptions of images matching the high-level semantic content in most of cases. Sometimes, captions are more general with descriptions like "animals in the grass" instead of the specific type of animal. In other cases, only details are missing (or wrong). For example, in Fig \ref{fig:captions} for the surfer image for one subject, the model adds "on a wave" while for the other the model specifies "on a beach". Similarly, in the first image of the right part, the pastry in the man's hand is changed to "a bowl of foods" or "slice of pizza". This could support the hypothesis that our pipeline is able to capture the main characteristic of the images from brain activity and the GIT decoder help in plausible sentence decoding.

\begin{table}[]
\resizebox{\textwidth}{!}{\begin{tabular}{llllllll}
              & \textbf{PixCorr} & \textbf{SSIM} & \textbf{AlexNet (2)} & \textbf{AlexNet (5)} & \textbf{Inception} & \textbf{CLIP}  \\
Lin et al     & -                & -             & -                    & -                    & 0,782              & -                          \\
Takagi et al  & -                & -             & 0,83                 & 0,83                 & 0,76               & 0,77                       \\
Gu et al      & 0,15             & 0,325         & -                    & -                    & -                  & -                            \\
Ozcelik et al & 0,30           & 0,28         & \textbf{0,89}                & \textbf{0,98}                & \textbf{0,92}              & \textbf{0,94}               \\
\textbf{Ours} & \textbf{0,353}   & \textbf{0,327}         & \textbf{0,89}                & 0,97       & 0,84     & 0,90        
\end{tabular}}

\caption{Image metrics. Metrics from Ozcelik et al were recomputed by requesting images from subj01 and subj02 to authors and averaged to better compare with our results, while metrics from other works are taken from original articles.}
\label{tab:img}
\end{table}

\begin{figure}[h!]
\centering
\includegraphics[width=0.97\textwidth]{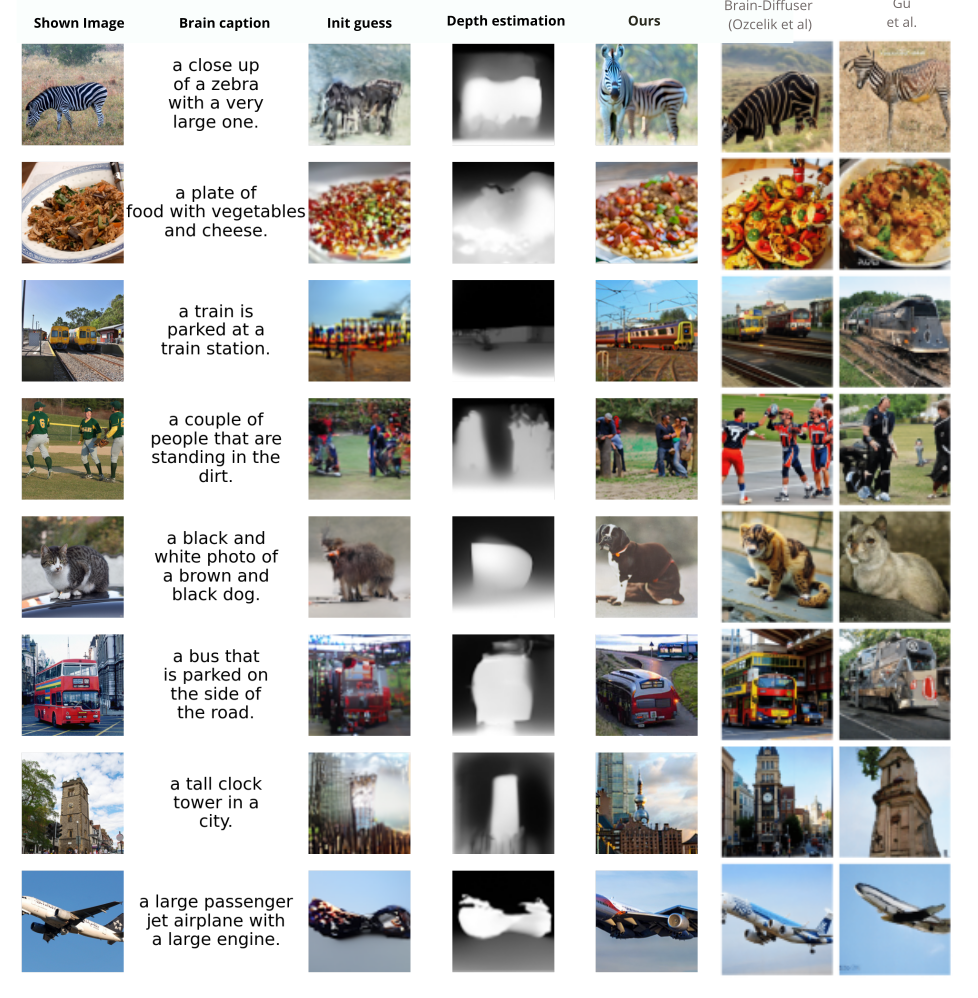}
\caption{Comparison of our results (Columns 2-4) with the presented stimuli and other reconstruction works. The second column displays the caption derived from brain activity, the third column presents the initial guess image, the fourth column exhibits the depth-estimated images, and the fifth column showcases our final reconstruction. The last two columns demonstrate reconstructions from two recent works. All results are from subj01.}
\label{fig:compare_gu}
\end{figure}

\vspace{-3mm}

\section{Discussion}

In this study, we proposed a method to generate captions from brain activity measured during a vision task. The primary motivation for shifting from image reconstruction to image captions is the flexibility of manipulating text prompts and the ease of modifying the image reconstruction pipeline as separate modules.
We also proposed an image reconstruction pipeline that incorporates depth maps and initial guesses to generate plausible images. Depth maps provide information about the spatial relationships between objects in a scene injecting information that could improve the overall quality of the reconstructed images.

\textbf{Neural Art and Examples}: Our approach has potential applications in neural art and style transfer. By leveraging our image reconstruction pipeline, we can explore the creative space of combining content and style from different text prompts. This could lead to the generation of visually captivating art, expanding the possibilities for artistic expression using AI. For example, modifying inputs by adding specific styles could drive the diffusion process toward an image with the same content but a different style. This approach represents a novel type of art that combines artificial intelligence, neuroscience, and creativity, starting from the decoded activity of the brain that could be modulated by a text description of the scene. For instance, it is recognized that attention can alter the semantic representation in the visual cortex. In the case of artists, they can undergo scans and deliberately focus on different aspects or mentally overlay additional details onto the observed images. Consequently, the resulting top-down regulations in the visual cortex are likely to modify the representation of the images, subsequently impacting the decoding process. This could potentially unveil distinctive artistic styles or other intricate details that are reflective of the artist's creative mind.

\textbf{Ethics}: As brain decoding research advances, ethical considerations must be addressed. For instance, the potential misuse of image reconstruction and generative models to create misleading or harmful content raises concerns, given that decoded activity is related to the mental and internal states of someone. It is crucial to develop guidelines and policies that ensure responsible use and prevent the exploitation of this technology for malicious purposes. Additionally, we must consider potential biases in the training data, as these can propagate and influence the generated output, perpetuating stereotypes and unfair representations, unrelated to thoughts of the specific subject. There are also possible concerns about privacy, given that brain decoding models are able to decode language, thoughts, and perceptions \citep{CEBRA,tang_semantic_2023_semantic_language}. From early experiments, it seems that high-level performances are only achievable when subjects are collaborating because the attention process can warp \citep{cukur_attention_2013} the semantic representation in the brain, which is the primary target of these deep learning multimodal models used as a proxy for brain activity \citep{choksi_multimodal_2022}.

\textbf{Portable Devices and Applications}: To overcome the limitations of subject-specific models, future research could project fMRI activity onto a common space to create a single inter-subject decoding model for fast transfer learning. Our flexible approach enables integration with portable devices for real-time image reconstruction, transitioning from fMRI to fNIRS\citep{fnirs}, which could impact various industries and enhance user experience.

\textbf{Limitations}

In our investigation of brain decoding, we have identified several key limitations that impact the efficacy and generalizability of our findings. The following discussion aims to elaborate on these constraints, establish their interconnections, and provide a deeper understanding of the challenges we face in advancing this field of research.
A major limitation in brain decoding work is the necessity for subject-specific models. Individual differences in brain structure, function, and cognitive processing make it challenging to develop a universal decoding model. This specificity hinders the broader applicability of our findings and demands the development of personalized models for each subject. Furthermore, it complicates cross-subject analyses and potentially limits the overall progress in the field. This could be an active area of future research, based on research of common functional spaces across subjects. \\
Even for subject-specific models, to achieve reliable and accurate decoding, a significant amount of high-quality data is needed. Obtaining such data is often time-consuming and resource-intensive, limiting the scalability of brain decoding studies. Additionally, low SNR data can introduce errors and inconsistencies in the decoding process, further compromising the reliability of the results. In this work we used a 7T dataset, that inherently has higher SNR with respect to previous 3T datasets \citep{horikawa_generic_2017}, enhancing the quality of our results.
In our work, the image captioning model acts as an upper limit: the performance of our brain captioning pipeline is inherently limited by the GIT image captioning model employed. Any inaccuracies or biases present in the model will directly impact the quality of decoded information, setting an upper bound on the performance that can be achieved. Improvements in the image captioning model and brain to features mapping are essential for enhancing the fidelity and scope of our decoding results, so our approach could be updated as new image captioning models will come out.
Also, the quality of the mapping between neural activity and external stimuli representation in latent spaces is another critical factor influencing the performance of our approach. This determines the accuracy and resolution of the decoded information. Current methods, however, are often limited by the complexity and variability of brain activity, as well as the constraints imposed by the data acquisition techniques, and usually rely on simple regression techniques. Addressing these challenges is essential for refining the mapping process and improving decoding outcomes.
Regarding image reconstruction, generating images from text could be another bottleneck. If the text contains errors, these will be propagated and/or enhanced by a separate image reconstruction pipeline. This represents the price for increased flexibility and independence from the specific image reconstruction pipeline used.
Finally, the brain decoding process may involve multiple areas, including temporal poles, which further impact of performances. Different brain regions may process and represent information differently, and understanding these variations is crucial for developing accurate and comprehensive decoding models. With the aim of reducing spatial dimensionality, we used only a visual responding region defined by the NSDGeneral ROI, however other brain areas could also encode relevant pieces of information that are relevant to improve performances. Exploring performances as a function of different input regions could be an interesting field of future research. 

\section{Conclusions}

Our approach builds upon neuroscientific and AI concepts, leveraging multimodal models to generate captions from brain activity related to the vision of different scenes. We augmented our brain captioning with a pipeline for image reconstruction that uses predicted text and initial information about colors and depth also estimated by brain activity.
In conclusion, our approach demonstrates promising results in image captioning and reconstruction from brain activity, with potential applications in a number of cross-disciplinary fields.
By drawing on these foundations, we could  further our understanding of the human brain's processing of visual and language information, ultimately improving related AI algorithms as well as applications. As we refine our approach, we can continue to explore the intricate relationship between neuroscience and AI, potentially uncovering novel insights and fostering interdisciplinary collaboration.

\bibliographystyle{plainnat}
\bibliography{ref}

\newpage
\section{Supplementary Material}
\renewcommand{\thefigure}{A\arabic{figure}}

\setcounter{figure}{0}

In this section, more comparisons of captions and reconstructed images are provided, compared with state-of-the-art brain decoding pipelines.

\begin{figure}[h!]
\centering
\includegraphics[width=1\textwidth]{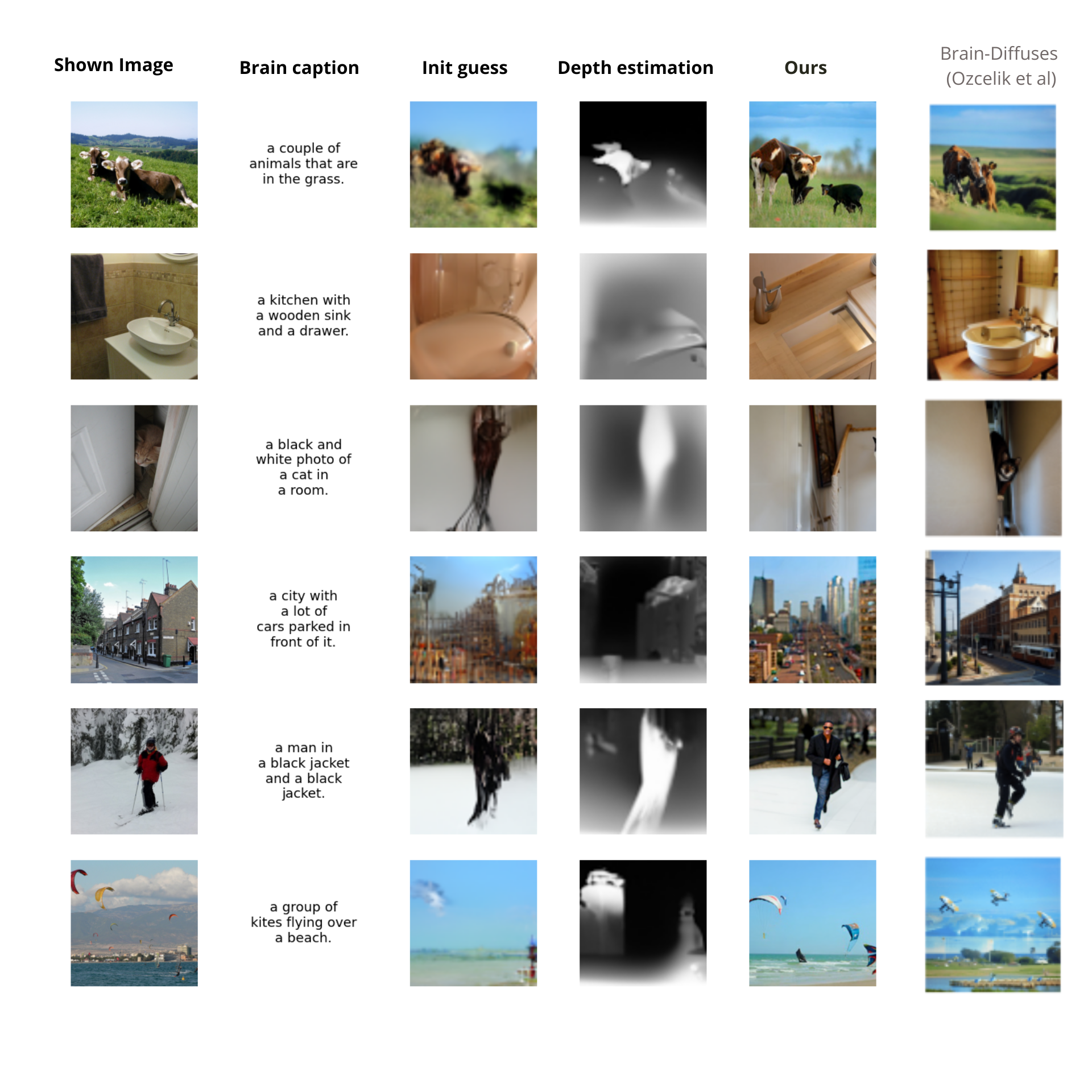}
\caption{Comparison of our results (Columns 2-4) with the presented stimuli and other reconstruction works. The second column displays the caption derived from brain activity, the third column presents the initial guess image, the fourth column exhibits the depth-estimated images, and the fifth column showcases our final reconstruction. The last column demonstrates reconstructions from the recent BrainDiffuser work. All results are from subj01.}
\label{fig:comparison_A1}
\end{figure}

\begin{figure}[h!]
\centering
\includegraphics[width=1\textwidth]{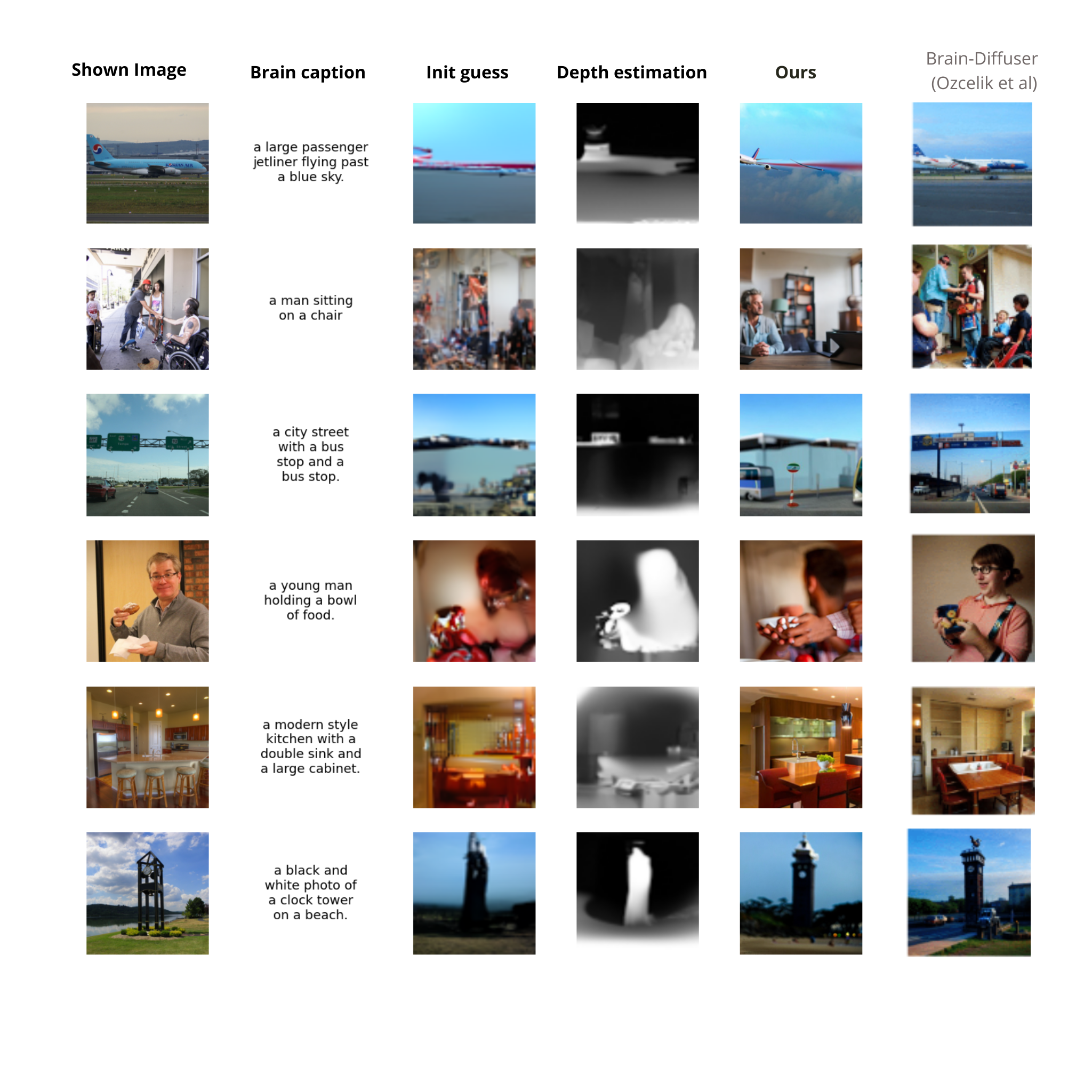}
\caption{Comparison of our results (Columns 2-4) with the presented stimuli and other reconstruction works. The second column displays the caption derived from brain activity, the third column presents the initial guess image, the fourth column exhibits the depth-estimated images, and the fifth column showcases our final reconstruction. The last column demonstrates reconstructions from the recent BrainDiffuser work. All results are from subj01.}
\label{fig:comparison_A2}
\end{figure}

\end{document}